\documentclass[10pt,twocolumn,letterpaper]{article}

\usepackage[pagenumbers]{cvpr} % To force page numbers, e.g. for an arXiv version

\usepackage{graphicx}
\usepackage{amsmath}
\usepackage{amssymb}
\usepackage{booktabs}
\usepackage{enumitem}

\usepackage[export]{adjustbox}
\usepackage{xspace}

\makeatletter
\DeclareRobustCommand\onedot{\futurelet\@let@token\@onedot}
\def\@onedot{\ifx\@let@token.\else.\null\fi\xspace}

\def\vs{vs\onedot}

\makeatother

\usepackage[pagebackref,breaklinks,colorlinks]{hyperref}

\usepackage[capitalize]{cleveref}
\crefname{section}{Sec.}{Secs.}
\Crefname{section}{Section}{Sections}
\Crefname{table}{Table}{Tables}
\crefname{table}{Tab.}{Tabs.}

\def\cvprPaperID{*****} % *** Enter the CVPR Paper ID here
\def\confName{CVPR}
\def\confYear{2022}

\begin{document}

\title{BEHAVIOR in Habitat 2.0: Simulator-Independent Logical Task Description for Benchmarking Embodied AI Agents}

\author{Ziang Liu$^{1}$, Roberto Martín-Martín$^{1}$, Fei Xia$^{2}$, Jiajun Wu$^{1}$, Li Fei-Fei$^{1}$\\
$^{1}$Stanford University \: $^{2}$Google Research\\
{\tt\small \{ziangliu, robertom, feixia, jiajunwu, feifeili\}@cs.stanford.edu}
}
\maketitle

%%%%%%%%% ABSTRACT
% \begin{abstract}
%   TODO
% \end{abstract}

%%%%%%%%% BODY TEXT
\section{Introduction}
\label{sec:intro}
% \vspace{-0.1cm}
Robots excel in performing repetitive and precision-sensitive tasks in controlled environments such as warehouses and factories~\cite{census}. However, this excellence has not been yet extended to embodied AI agents providing assistance in uncontrolled environments, i.e. assisting humans in everyday tasks at home. Inspired by the catalyzing effect that benchmarks have played in the AI fields such as computer vision~\cite{imagenet,kitti} and natural language processing~\cite{glue}, the community is looking for new benchmarks for embodied AI, often running on simulation to leverage their safety, reproducibility and speed. Different to the very clear and uniformly adopted definitions of success in CV and NLP benchmarks, in embodied AI each benchmark defines tasks using a different formalism, often specific to one environment, simulator or domain, making it hard to develop general and comparable solutions.

The most common ways to define embodied AI tasks include geometric, image, language, experience, and predicate~\cite{rearrangement}. The most adopted is geometry: manually defining regions to place objects (rearrangement) or the robot (navigation) to claim success~\cite{habitat,tdw, metaworld, robosuite, rlbench}. While providing an exact guidance to the agent, this formulation requires explicit knowledge of the object/robot valid goal poses for each scene, involving a manual process that does not generalize to other scenes. Experience-based goal allow agents to collect observations in the goal environment~\cite{ai2thor, rearrangement}. However, despite its simplicity, providing a goal environment in the real world is challenging. Language goal describe configurations with natural language~\cite{alfred}. This formulation is the closest to defining in the logic domain and more interpretable to human, but is less concise and adds the challenge of language understanding. To alleviate the aforementioned limitations, we note that using logic predicates to define tasks provides more generalizability to different scenes and simulators, and is closer to real world task definitions.

BEHAVIOR~\cite{behavior} is a set of 100 household activities for evaluating embodied AI agents defined in BEHAVIOR Domain Definition Language (BDDL) with synsets and logic predicates instead of grounded object instances and representations that depend on simulator features, providing a level of abstraction that can be adapted to any simulator and scene assets while allowing a flexible configuration space similar to how humans define tasks in the real world. Although BEHAVIOR is simulator-agnostic, so far it has only been integrated with iGibson 2.0 (iG 2.0)~\cite{igibson}. Recent release of Habitat 2.0 (H2.0)~\cite{habitat} shows a promising test bed for BEHAVIOR, as they provide a significantly higher simulation speed and thus allowing more experiences in the same time period. 

In this work, we bring 45 out of the 100 BEHAVIOR activities which involve only kinematic states into H2.0 to benefit from its fast simulation speed as a first step towards demonstrating the ease of adapting activities defined in the logic space into different simulators, in the process equip H2.0 with a even richer set of iG 2.0 interactive scenes and assets.

\begin{figure*}[]
  \centering
  \adjincludegraphics[width=\linewidth, trim={{0.07\width} {0.54\height} {0.09\width} {0.09\height}}, clip]{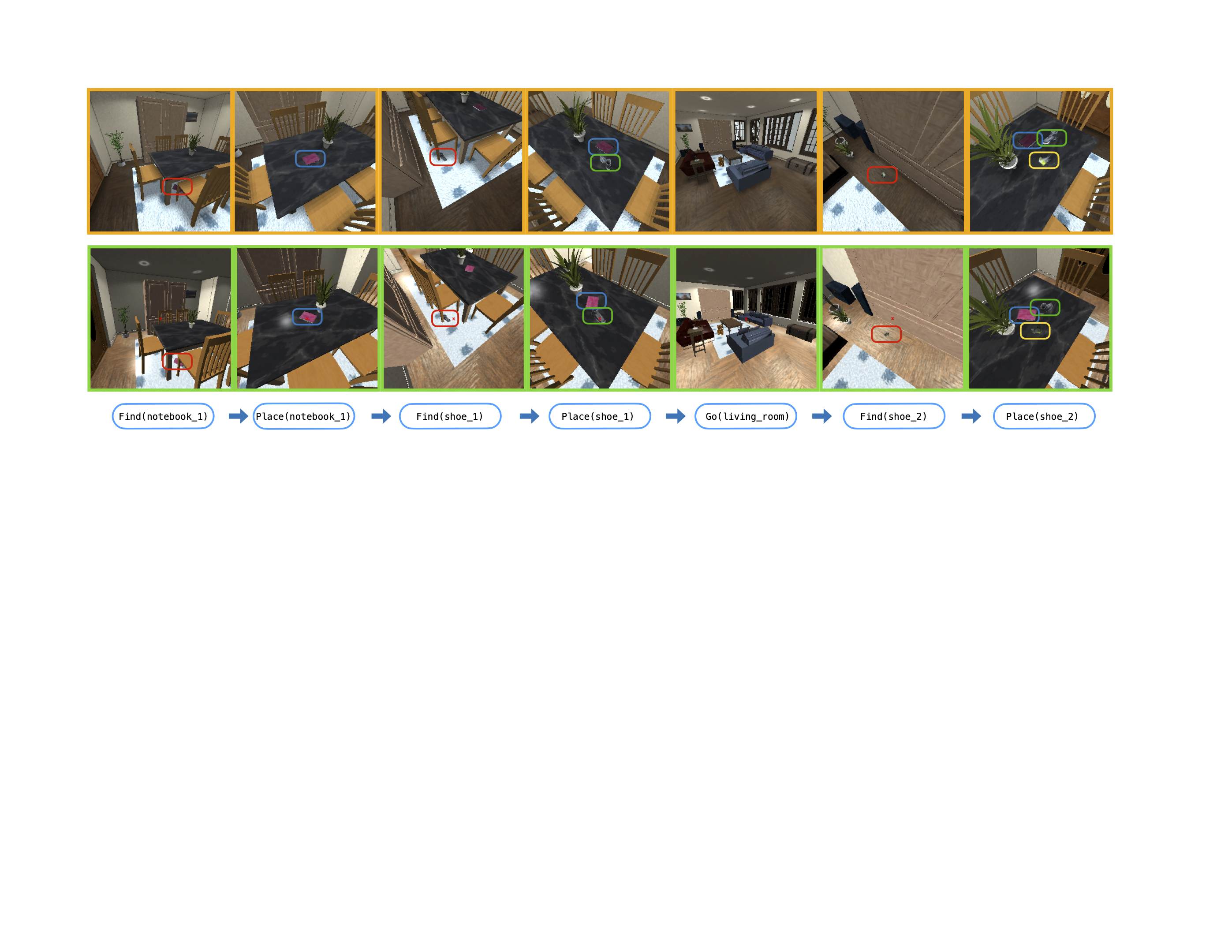}
  \vspace{-0.5cm}
  \caption{Performing one episode of BEHAVIOR activity to {\em collect\_misplaced\_item} through teleoperation. Top row: observation key-frames from iG 2.0. Bottom row: observation key-frames from H2.0.}
  \label{fig:teleop}
% \vspace{-0.2cm}
\end{figure*}
%-------------------------------------------------------------------------
\section{BEHAVIOR in H2.0}
% \vspace{-0.1cm}
Fully supporting BEHAVIOR activities in a new simulator imposes five requirements, as stated in Section 5 in the BEHAVIOR paper~\cite{behavior}: 1) object-centric representation, 2) simulate physics and sensor signals, 3) non-kinematic states, 4) instance sampling from BDDL conditions, 5) state predicate checking. H2.0 naturally satisfies requirement 2) through its variety of sensor signals and the physics simulation through Bullet~\cite{bullet}. In this work, we extend H2.0 for the requirements 1), 4) and 5). For 1), we extend the H2.0 simulator to keep track of the additional object-centric state information needed for evaluating activity progress with BDDL. For 4), we enable H2.0 to use iG 2.0 assets and leverage the sampled instances from iG 2.0. For 5), we implement the pipeline to evaluate each kinematic state predicate. The missing requirement 3) is a current limitation of our effort: we have only kinematic states. This restricts our effort to support 45 out of 100 BEHAVIOR activities.

\vspace{-0.4cm}
\paragraph{Loading BEHAVIOR Instances in H2.0.}

Many objects in daily household activities require interaction with objects' articulation mechanisms, from loading dishes into a dish washer to opening doors and beyond. For simulators to support scenes that closely resemble real world scenarios, having more articulated objects that represents their real world counterparts in various scene layouts is highly desirable. H2.0's ReplicaCAD dataset lacks abundance in object categories, articulated objects, and scene layouts, despite the richness in carefully designed room configurations, as shown in Table \ref{tab:ig_vs_hab_asset}.  Adding iG 2.0 scenes and assets allows H2.0 users to train and evaluate their AI agents with far more diverse environments and object set, and in particular more articulated objects to interact with.

\vspace{-0.5cm}

\paragraph{Checking Predicates for BEHAVIOR Activities in H2.0.}

BEHAVIOR requires seven kinematic states (NextTo, OnTop, etc.) and fourteen non-kinematic states (Burnt, Sliced, etc.). In this work we focus on implementing kinematic states in H2.0 that are essential for many BEHAVIOR activities. We provide a BDDL backend for H2.0 that supports predicate checking for {\em NextTo} , {\em Inside}, {\em OnFloor}, {\em OnTop}, {\em Touching}, and {\em Under}. For validating task completion progress and task success, we leverage the logic evaluation mechanism from BDDL. Overall, our effort facilitates training and evaluating on 45 out of 100 BEHAVIOR activities.

%-------------------------------------------------------------------------
\section{Experimental Validation}
\vspace{-0.1cm}
To demonstrate and validate our implementation, we perform an episode of the {\em collect\_misplaced\_item} activity in the {\em Wainscott\_0\_int} apartment in both iG 2.0 and H2.0 through teleoperation. 

The captured key-frames in Figure \ref{fig:teleop} correspond to observations when performing the activity. Benefited from BEHAVIOR's logic domain specification, we are able to implement the same activities in two different simulators without altering the activity definition in any way. Note that the differences in object appearances are due to lighting setup and using non-pbr rendering in H2.0.

\vspace{-0.5cm}
\paragraph{Performance Comparison of iG 2.0 \vs H2.0.} 
One of our goals in bringing BEHAVIOR to H2.0 is to gain performance benefit. Our effort enables a fair performance comparison of iG 2.0 and H2.0 with the same assets.

From our evaluation, H2.0 provided 10.4x speed up in an iG 2.0 scene with 64 processes on 8 GPUs. However, as the number of objects increase, the performance benefit of H2.0 over iG 2.0 decreases to 1.5x with 16 processes on 1 GPU and 1.25x with 64 processes on 8 GPUs, and 0.94x on a single process.
 
\begin{table}
  \centering
  \begin{tabular}{c c c c c c}
    \toprule
    Asset & Apt. & Rm. & Cat. & Obj. & A.O. \\
    \midrule
    BEHAVIOR & \textbf{15} & \textbf{100} & \textbf{391} & \textbf{1217} & \textbf{339}\\
    ReplicaCAD & 1 & 1 & 41 & 1201 & 8\\
    \bottomrule
  \end{tabular}
  \caption{BEHAVIOR (iG 2.0) and ReplicaCAD (H2.0) assets comparison, based on the number of apartments, rooms, layouts, object categories, objects, and articulated objects.}

  \label{tab:ig_vs_hab_asset}
  \vspace{-0.5cm}
\end{table}

%-------------------------------------------------------------------------
\vspace{-0.1cm}
\section{Next Steps}
\vspace{-0.1cm}
In this work, we ported BEHAVIOR household activities into H2.0, demonstrating that defining tasks in the high-level logic domain allows simple implementation of the tasks in different simulators. To further demonstrate the behavior of agents trained on the same task in different simulators, we plan to provide simple baseline training results in both iG 2.0 and H2.0, and release the code publicly to facilitate research in this direction. As a main limitation, our work currently only enabled activities with kinematic states; a natural extension is to implement the relevant extended object states and predicate checking mechanisms for the non-kinematic states to support even more BEHAVIOR activities.

%%%%%%%%% REFERENCES
{\small
\bibliographystyle{ieee_fullname}
\bibliography{egbib}
}

\end{document}